\documentclass[11pt,a4paper]{article}
\usepackage[hyperref]{emnlp2020}
\usepackage{times}
\usepackage{latexsym}
\usepackage{multicol}
\usepackage{multirow}
\usepackage{tabu}
\usepackage{caption}
\usepackage{subcaption}
\usepackage{bm}
\usepackage{CJK}
\usepackage{amsmath}

\usepackage{microtype}

\usepackage{graphicx}
\aclfinalcopy 
\def\aclpaperid{***} 

\title{{C}ross {C}opy {N}etwork for {D}ialogue {G}eneration}

\author{Changzhen Ji$^1$, Xin Zhou$^2$, Yating Zhang$^2$\\
\textbf{Xiaozhong Liu$^3$, Changlong Sun$^2$, Conghui Zhu$^1$ and Tiejun Zhao$^1$}\\
$^1$Harbin Institute of Technology, Harbin, China\\
$^2$Alibaba Group, Hangzhou, China\\
$^3$Indiana University Bloomington, USA\\
{\small\tt czji\_hit@outlook.com, \{eric.zx,ranran.zyt\}@alibaba-inc.com}\\
{\small \tt liu237@indiana.edu, changlong.scl@taobao.com, \{conghui,tjzhao\}@hit.edu.cn}}

\date{}

\begin{document}
\maketitle
\begin{abstract}
In the past few years, audiences from different fields witness the achievements of sequence-to-sequence models (e.g., LSTM+attention, Pointer Generator Networks and Transformer) to enhance dialogue content generation.
While content fluency and accuracy often serve as the major indicators for model training, dialogue logics, carrying critical information for some particular domains, are often ignored. Take customer service and court debate dialogue as examples, compatible logics can be observed across different dialogue instances, and this information can provide vital evidence for utterance generation. 
In this paper, we propose a novel network architecture - Cross Copy Networks (\textbf{CCN}) to explore the current dialog context and similar dialogue instances’ logical structure simultaneously. Experiments with two tasks, court debate and customer service content generation, proved that the proposed algorithm is superior to existing state-of-art content generation models.
\end{abstract}

\maketitle

\section{Introduction}
\begin{figure*}
    \centering
    \includegraphics[width=1\textwidth]{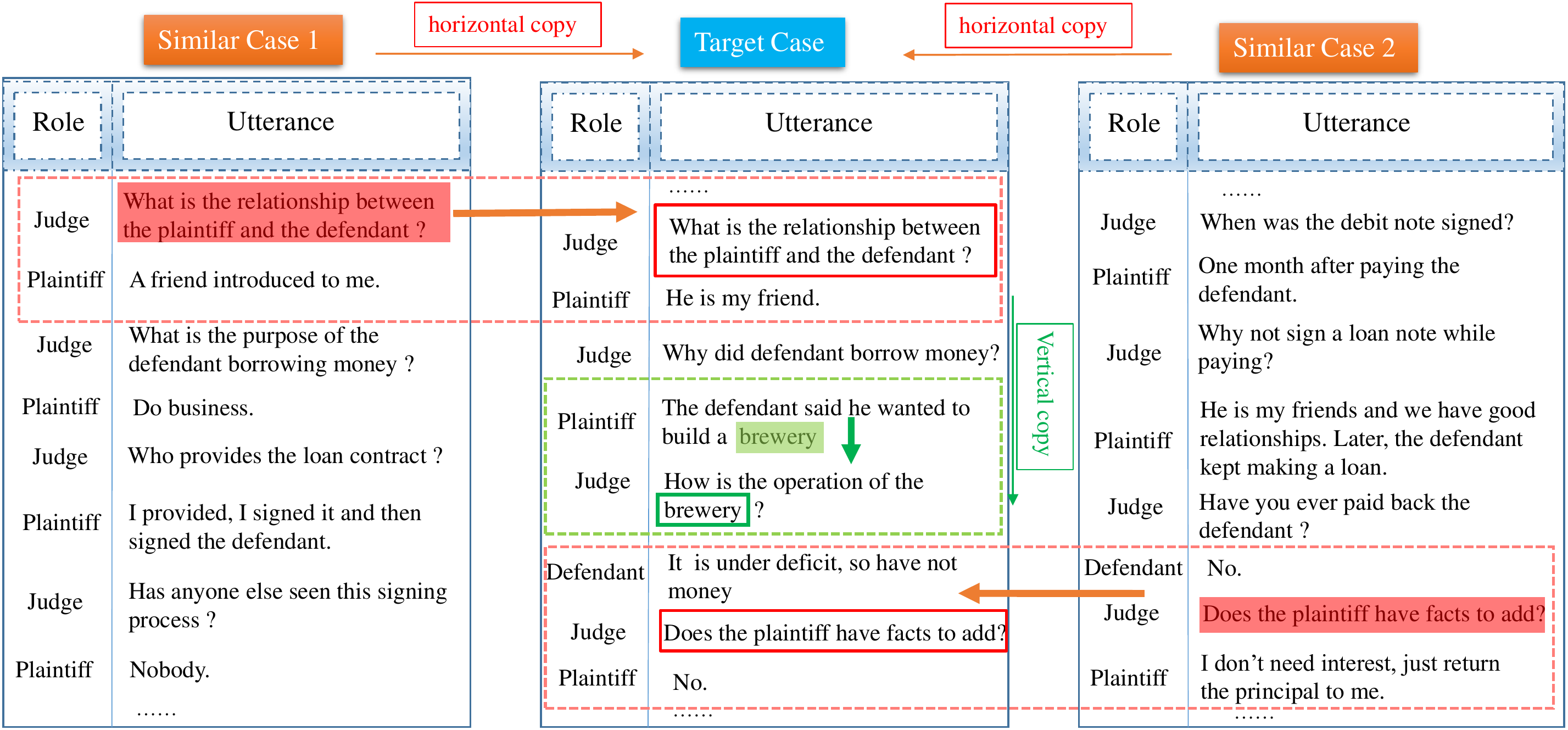}
    \caption{An example from the court debate showing the intuition of utterance generation by leveraging its context and the information from its similar cases. We name the copy process from the context as \textit{vertical copy} and the one copied from its neighbor cases as \textit{horizontal copy}.}
    \label{fig:similar}
\end{figure*}

As an important task in Natural Language Generation \textbf{(NLG)}, dialogue generation empowers a wide spectrum of applications, such as chatbot and customer service automation.
In the past few years, breakthroughs in dialogue generation technology focused on a series of sequence-to-sequence models \cite{sutskever2014sequence}. 
More recently, external knowledge is employed to enhance model performance.
\newcite{wu2019proactive, liu2018knowledge} can assist dialogue generation by using knowledge triples.
Similarly, \newcite{li2019incremental, rajpurkar2018know, huang2018flowqa, reddy2019coqa} explore document as knowledge discovery for dialogue generation, and
\newcite{xia2017deliberation, ye2020knowledge, ghazvininejad2018knowledge, parthasarathi2018extending} utilize unstructured knowledge to explore in the open-domain dialogue generation.
However, unaffordable knowledge construction and defective domain adaptation restrict their utilization. 

Copy-based generation models \cite{vinyals2015pointer,gu2016incorporating} have been widely adopted in content generation tasks and show better results compared to sequence-to-sequence models when faced with out-of-vocabulary problem. Thanks to their nature of leveraging vocabulary and context distributions for content copy, it enables to copy the aforementioned named entities (e.g., person names, locations, company names) appeared in the above context) from the upper context to improve the specificity of the generated text.  

In the task of dialogue generation, we can often observe the phrases/utterance patterns across different "similar dialogue" instances. For example, in customer service, the similar inquiries from the customers will get similar responses from the staff. It motivates us to build a model that can not only copy the content within the upper context of the target dialogue instance, but also learn the similar patterns across different similar cases of the target instance. Such external copy can be critical in some scenarios. 

As shown in Figure.\ref{fig:similar}, we propose two different kinds of copy mechanisms in this study: \textbf{vertical copy} context-dependent information within the target dialogue instance, and \textbf{horizontal copy} logic-dependent content across different 'Similar Cases' \textbf{(SC)}. This framework is labeled as Cross-Copy Networks \textbf{(CCN)}. As exemplar dialogue depicted, judges may repeat (horizontal copied) words, phrases or utterances from historical dialogues when those SCs sharing similar content, e.g., `\textit{A sue B because of X and Y}'.




In order to validate the proposed model, we employ two different dialogue datasets from two orthogonal domains - \textit{court debate} and \textit{customer service}. We apply proposed \textbf{CCN} to both datasets for dialogue generation. Experiments show that our model achieves the best results. To sum up, our contributions are as follows:

\begin{itemize}
\item We propose a new end-to-end model, the Cross Copy Networks $(\textbf{CCN})$, which enables internal (vertical) copy from the target dialogue and external (horizontal) copy from similar cases in the dataset without employing any external resources. 
\item We validate the proposed model by leveraging two different datasets - court debate and customer service datasets. Experiments show that our model has achieved State-of-the-art results in both domain datasets.
\item To motivate other scholars to investigate this novel but
an important problem, we make the experimental datasets publicly available\footnote{https://github.com/jichangzhen/CCN}.

\end{itemize}
\section{Model}
\begin{figure*}
    \centering
    \includegraphics[width=1\textwidth]{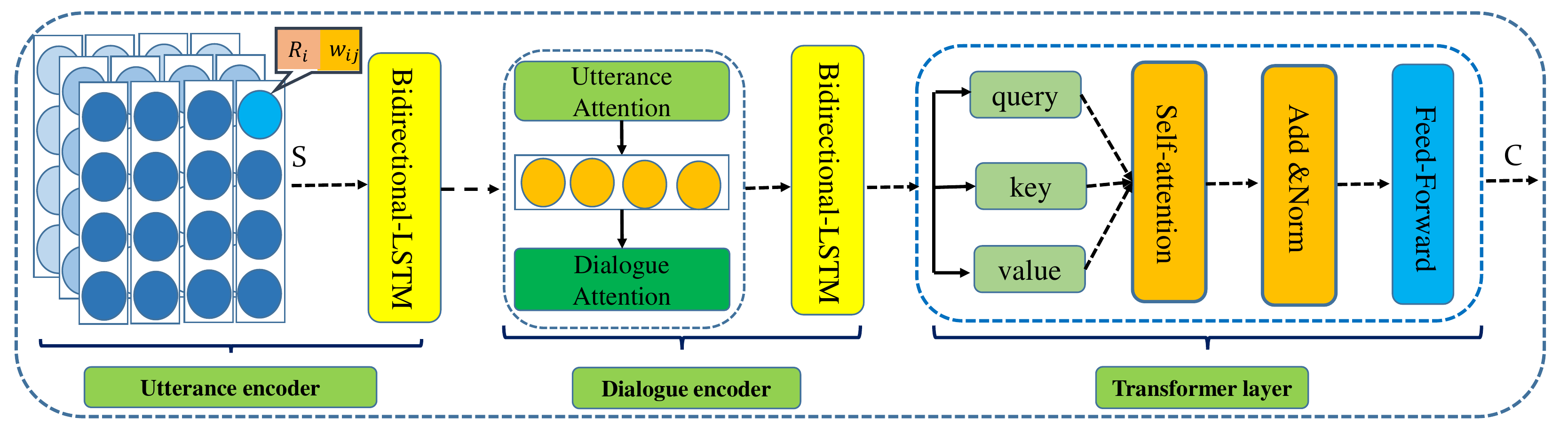}
    \caption{The encoder of \textbf{CCN} is divided into three levels: (1) \textbf{Utterance layer}: is used to encode role information and word level information; (2) \textbf{Dialogue layer}: is used to encode sentence level information; (3) \textbf{Transformer layer}: is used to capture long distance dependence for dialogue.
    }
    \label{fig:encoder}
\end{figure*}

In this section, we introduce the proposed model, the Cross Copy Network, which has three major components:

\begin{enumerate}
\item \textbf{Target Case Representation}: we obtain the target case representation with two attention distributions at the utterance layer and the dialogue layer, which contribute to the final attention distribution (Section 2.1);
\item \textbf{Similar Case Representation}: we fine-tune the pre-trained language model to obtain similar cases, and adopt the same method as the target case for encoding \textbf{SC} (Section 2.2);
\item \textbf{Cross Copy}: we learn two pointer distributions which are used to achieve internal (vertical) copy and external (horizontal) copy respectively (Section 2.3).
\end{enumerate}

\subsection{Target Case Representation}
Given a dialogue $\mathbf{D}=\{(\mathbf{U}, \mathbf{R})^L\}$ containing $L$ utterances, the $\mathbf{U}$ and $\mathbf{R}$ stand for utterance and role of speaker, respectively, where each utterances in the dialogue is expressed as $\mathbf{U_i}=\{w_{i1}, w_{i2},...,w_{il}\}$, and the $l$ represents the length of the utterances. To distinguish \textbf{SC} and original context, we define the original context (historical dialogue) as \textbf{Target Case}.


Our encoder is shown in Figure \ref{fig:encoder}. It is designed with hierarchical infrastructure consisting of three levels of components: utterance layer, dialogue layer and transformer layer.

\subsubsection{Utterance Layer}
In the dialogue, role information can make critical contribution to the task of dialogue generation, and different roles may not share consistent lexical spaces. For role information $\mathbf{R_i}$, we utilize a $100$-dimensional vector to represent different roles which is randomly initialized, and updated via back propagation. 

To take the role information into consideration for utterance representation learning, we concatenate the role information with each word of utterance expressed as $\mathbf{S_{ij}}$, and we use Bidirectional Long-Short Term Memory networks \textbf{(Bi-LSTM)} \cite{hochreiter1997long} to encode the semantics of the utterance while maintaining its syntactic be expressed as $\mathbf{h^d}$.

In order to obtain the different importance of different historical dialogue information, we adopt the attention mechanism \cite{bahdanau2014neural} to obtain the utterance level's attention distribution in historical dialogue $a_j^u$ and utterance context $\mathbf{h_i^U}$: 
\begin{equation}
\mathbf{h_i^U}=\sum_{j=1}^{l}a_j^u\mathbf{h^d_{ij}}
\end{equation}
\begin{equation}
a_j^u = {\frac{exp(tanh(W^u\mathbf{h^d_{ij}}+b^u)^{\mathrm{T}}\mathbf{h^d_{ij}})} 
{\sum\nolimits_{j=1}^lexp(tanh(W^u\mathbf{h^d_{ij}}+b^u)^{\mathrm{T}}\mathbf{h^d_{ij}})}}
\end{equation}
The $a_j^u$ represents the word probability distribution for the target utterance\footnote{The $W^u$ and $b^u$ are learnable parameters, the $tanh$ is hyperbolic tangent function.}.

\subsubsection{Dialogue Layer}
In order to represent the context information of the dialogue, we also use \textbf{Bi-LSTM} to encode the utterance dependency to obtain a global representation of an utterance as dialogue, denoted as $\mathbf{h^D}$. 

We obtain the dialogue layer attention distribution $a_i^d$, which is a probability distribution over the prior utterances in the target dialogue. The $a_i^d$ can be expressed as: 
\begin{equation}
a_i^d = {\frac{exp(tanh(W^d\mathbf{h^D_{i}}+b^d)^{\mathrm{T}}\mathbf{h^D_{i}})} 
{\sum\nolimits_{i=1}^Lexp(tanh(W^d\mathbf{h^D_{i}}+b^d)^{\mathrm{T}}\mathbf{h^D_{i}})}}
\end{equation}
The final context attention distribution $A^d$ of target case can be expressed as the product of $a_j^u$ and $a_i^d$:
\begin{equation}
A^d = a_j^u * a_i^d
\end{equation}
\subsubsection{Transformer Layer}
To expand the model's ability to focus on different locations of long context, we adopt the self-attention with multi-heads \cite{vaswani2017attention} to explore an enhance representation, denoted as Transformer-Block.
We feed $\mathbf{h^D}$ to a N-layer Transformer-Block to suppress the long distance dependency for dialogue. Following this strategy, the final target case representation is:
\begin{equation}
\mathbf{C^d}=\mathbf{Transformer^N{(h^D)}}
\end{equation}

\begin{figure*}
    \centering
    \includegraphics[width=1\textwidth]{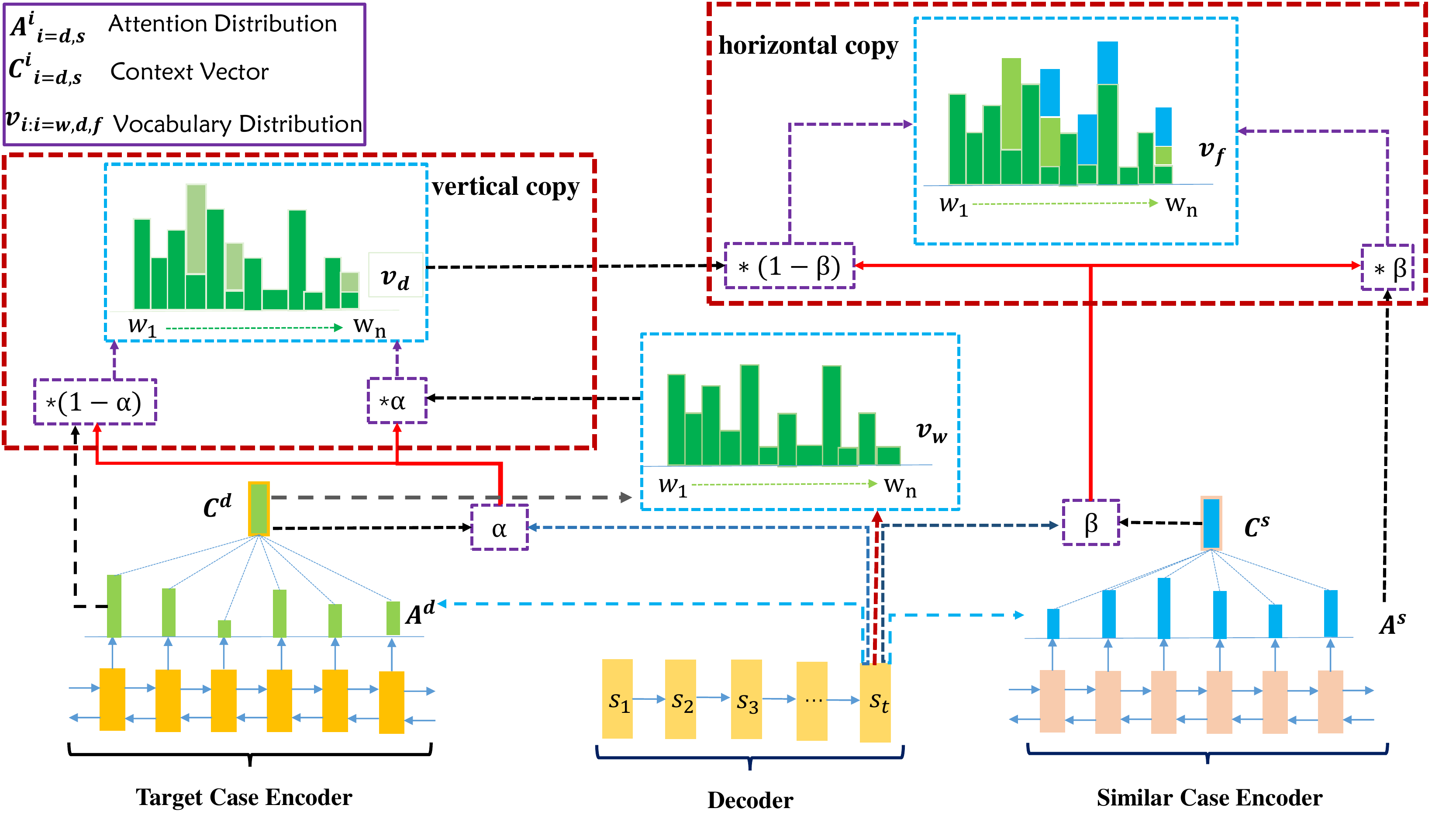}
    \caption{The decoder of \textbf{CCN} learns two pointer distribution to extend the original vocabulary twice. It learns the pointer distribution $\alpha$ to obtain the content to be copied from the context (vertical copy) as well as the pointer distribution $\beta$ to determine the content to be copied from its similar cases (horizontal copy).}
    \label{fig:decoder}
\end{figure*}

\subsection{Similar Case Representation}
In this section, we introduce the approach of obtaining and representing similar cases.

\subsubsection{Similar Case Finding}
The similar cases $(\textbf{SC}s)$ of the target case is discovered from the same dataset where the target case stays. To make it more efficient, we use ElasticSearch\footnote{\url{https://www.elastic.co/products/elasticsearch}} to retrieve top $50$ similar cases as candidates by leveraging the target case as a query and the all the other cases as documents. To make it more effective, we fine-tune the pre-trained \textbf{RoBERTa}\footnote{All the dialogues in the dataset are used to fine-tune RoBERTa. The average lengths of the candidate cases and the target case are both 106, so it can all fit within the RoBERTa model} \cite{liu2019roberta} model. It uses a linear layer with sigmoid activation function on top of the pooled [CLS] representation from the concatenation of the target case and each candidate retrieved above as a binary classifier, to obtain similarity score.


\subsubsection{Similar Case Encoding}
For \textbf{SC} encoding, we adopt the same method as the target case\footnote{We use two identical encoders to encode target case and similar case, while two encoders' parameters are not shared}.
We fuse role information with each word of utterance in the \textbf{SC} and then use \textbf{Bi\_LSTM} to obtain hidden state $\mathbf{h^s}$.
Next, we adopt the attention mechanism to obtain the utterance layer distribution and dialogue layer distribution.

Therefore, we obtain attention distribution for different words $a_j^{u*}$ of each utterance and different utterances $a_i^{d*}$ of each \textbf{SC}. 
Then, we get final attention distribution $A^s$ which can be expressed as the product of $a_j^{u*}$ and $a_i^{d*}$.

Finally, we use the N-layer transformer-block to obtain the final \textbf{SC} representation $\mathbf{C^s}$.

\subsection{Cross Copy}
In this section, we learn two pointers distribution and to achieve internal (vertical) copy and external (horizontal) copy. The decoder's structure shown as Figure \ref{fig:decoder}.

On the time step $t$, we concatenate target case context vector $\mathbf{C^d_t}$ with decoder states $\mathbf{s_t}$ to get the distribution of the current vocabulary:
\begin{equation}
v_{w} = softmax(W^v(W^w[\mathbf{C^d_t},\mathbf{s_t}]+b^w)+b^v)
\end{equation}
where the $W^w$, $b^w$, $W^v$ and $b^v$ are learnable parameters.

For the cross copy, the algorithm execution process is divided into two stages.

At the first stage, we perform \textbf{vertical copy}.
With the target case encoder hidden state $\mathbf{h^d}$, context vector $\mathbf{C^d}$, and decoder hidden states $\mathbf{s}$ as mentioned above, on the time step $t$, we can learn \textbf{vertical copy} probability distribution $\alpha$, it determines whether to copy the words from the historical dialogue. It can be expressed as Eq.\ref{Eq:alpa}:
\begin{equation}
\label{Eq:alpa}
    \alpha = \sigma(W^h*\mathbf{h^d_t}+W^c*\mathbf{C^d}_t+W^s*\mathbf{s_t}+b^d)
\end{equation}

Combined with the attention distribution $A^d$, we can get the dynamic extended vocabulary $v_d$ by pointer distribution $\alpha$:
\begin{equation}
v_d = \alpha*v_w+(1-\alpha)*\sum_{i:w_{i}=w}^{l*L}A^d
\end{equation}

In the second stage, we learn the \textbf{horizontal copy} probability distribution $\beta$ of \textbf{SC}. For \textbf{SC} context vector $\mathbf{C^s}$ and hidden state $\mathbf{h^s}$, on the time step $t$, we combine decoder hidden state $\mathbf{s}$, to get the \textbf{horizontal copy} pointer distribution $\beta$:
\begin{equation}
\beta = \sigma(W^{h*}*\mathbf{h^s_t}+W^{c*}*\mathbf{C^s_t}+W^{s*}*\mathbf{s_t}+b^s)
\end{equation}

From the encoder, we get the attention distribution $A^s$ of \textbf{SC}, we then perform a second expansion of the dynamic vocabulary to obtain the final vocabulary $v_f$:
\begin{equation}
v_f = (1-\beta)*v_d+\beta*\sum_{i:w_{i}=w}^{l*L}A^s
\end{equation}

It should be noted, if $w$ is not in original vocabulary but in \textbf{SC}, the $\beta$ is $1$. Model will copy the $w$ from \textbf{SC}. One of the advantages of our model is that it can produce  out-of-vocabulary word.

In the formula above, the $\sigma$ is sigmoid function. The $W^h$, $W^c$, $W^s$, $b^d$,$W^{h*}$, $W^{c*}$, $W^{s*}$ and $b^s$ are learnable parameters.

\subsection{Loss function}
In this dialogue generation task, for each dialogue $\mathbf{D}$, the loss function is defined as:
\begin{align*}
& loss = -\log P(\mathbf{S}\mid \mathbf{D}) \\
& \quad \  \, =-\sum_{j=1}^{l}\log P(w_{ij}\mid w_{i1:j-1},\mathbf{D})
\end{align*}

Denoting all the parameters in our model as $\delta$, then we obtain the following optimized objective function:
\begin{equation}
\underset{\theta}{min}\; loss = loss + \lambda\left \| \delta \right \|_2^2
\end{equation}

To minimize the objective function, we use the diagonal variant of Adam \cite{zeiler2012adadelta}. 

\section{Experimental Settings}

\begin{table}
\small
  \caption{Statistics of the CDD and JDDC}
  \label{tab:dataset}
  \begin{tabular}{c|c|c|c|c}
  \hline
    \multicolumn{1}{c|}{\multirow{2}{*}{\textbf{Dataset}}} &
    \multicolumn{2}{c|}{\textbf{CCD}} &
    \multicolumn{2}{c}{\textbf{JDDC}} \\ \cline{2-5}
     & Dialogue & Utterance & Dialogue & Utterance\\
    \hline
    train & 208,152 & 2,869,794  & 261,282 & 3,135,377\\
    \hline
    dev &  26,018 & 364,345 &  32,660 & 391,983\\
    \hline
    test & 26,020 & 371,554 & 32,661 & 391,480\\ 
    \hline
    \textbf{Total} & \textbf{260,190} & \textbf{3,605,693} 
    & \textbf{326,603} & \textbf{3,918,840}\\
    \hline
\end{tabular}
\end{table}

\subsection{Dataset}
We employed two datasets for the experiment, Court Debate Dataset \textbf{(CDD)} from judicial field and Jing Dong Dialogue Corpus \textbf{(JDDC)} \cite{chen2019jddc} from e-commerce field. The details of the dataset are illustrated in Table \ref{tab:dataset}. During training, the data is divided into the training set, development set and test set\footnote{The entire dataset is divided by a ratio of 8:1:1 for training, developing and testing, respectively.}.

\subsubsection{Court Debate Dataset}
For \textbf{CDD}, we collected $121,016$ court debate records of private lending dispute civil cases\footnote{Private lending dispute cases are the most frequent cause of civil cases in China. This data set is provided by the High People's Court of a province in China. All the court transcripts are manually recorded by the court clerk.}. 
We take the judge and the historical conversation with the plaintiff and the defendant as the model input, and the judge's utterance as the model output. These records are divided into $260,190$ pairs of samples by experts with legal knowledge.

\subsubsection{Jing Dong Dialogue Corpus}
Jing Dong Dialogue Corpus $\textbf{(JDDC)}$\footnote{\url{http://jddc.jd.com/auth\_environment}} contains $1,024,196$ multi-turn dialogues, $20,451,337$ utterances, and $150$ million words.
The average number of tokens contained in per sentence is about $7.4$.
In the experiments, we adopted the top $326,603$ cases. The proposed algorithm and baselines are set to generate the utterances of the customer service, and the historical context between the customer service and the customer is set as input.

\begin{table*}[!t]
\small
  \caption{Quantitative Evaluation. We report ROUGE-1, ROUGE-L and BLEU scores for each tested methods. }
  \vspace{-8pt}
  \label{tab:result_discussion}
  \begin{center}
  \begin{tabular}{c|c|c|c|c|c|c}
    \hline
    \multicolumn{1}{c|}{\multirow{2}{*}{\textbf{model}}} &
    \multicolumn{3}{c|}{\textbf{CCD}} &
    \multicolumn{3}{c}{\textbf{JDDC}} \\ 
    \cline{2-7}
    & \textbf{ROUGE-1} &  \textbf{ROUGE-L} & \textbf{BLEU} 
    & \textbf{ROUGE-1} &  \textbf{ROUGE-L} & \textbf{BLEU}\\
     \hline
    LSTM\cite{hochreiter1997long}           
    &  30.28 & 28.02  &  9.77 
    &  19.45 & 18.74  &  9.52  \\
    ByteNet\cite{gehring2017convolutional}        
    &  33.68 & 32.99  &  16.91
    & 22.19  & 18.35  &  11.55 \\
    ConvS2S\cite{kalchbrenner2016neural}       
    & 35.92 & 31.48  & 16.34      
    & 26.53 & 21.08  & 11.64 \\
    S2S+attention\cite{nallapati2016abstractive}  
    & 36.91 &  33.12  &  18.52
    & 28.44 &  22.34  &  13.42 \\
    PGN\cite{see2017get}            
    & 37.03  &  34.25  & 18.75   
    & 29.78  &  24.06  & 14.37 \\
    Transformer\cite{vaswani2017attention}    
    & 37.59  &  34.93  & 18.58
    & 27.25  &  22.75  & 11.29 \\
    CCN(vertical-only)
    & 37.15 &  34.51  &  19.65     
    & 30.27  & 25.08  &  15.75 \\
    \hline
    CCN$_{top-1}$
    &  39.12 & 39.23  & 23.11    	
    &  32.44 & 29.18 &  16.90 \\
    CCN$_{top-2}$
    & 40.43  & 38.16 &  23.24    
    & 33.56  & 31.17 &  18.52 \\
    CCN$_{top-3}$
    & $\mathbf{41.10}$  & $\mathbf{39.82}$  &  $\mathbf{24.75}$   
    & $\mathbf{34.17}$  & $\mathbf{32.37}$  &  $\mathbf{19.53}$ \\
  \hline
    \end{tabular}
    \vspace{-8pt}
\end{center} 
\end{table*}

\begin{table*}[!t]
\small
\centering
  \caption{Qualitative Evaluation. We report average score (Avg) and calculate the $\kappa$ value in relevance and fluency. We recruited five annotators to evaluate the sentences generated by all the models. To be fair, for each input, we shuffled the output generated by all the models and then let the annotator to evaluate. The $\kappa$ represents the consistency of evaluation by different annotators. And the $\kappa$ coefficient between 0.48 and 0.82 means middle and upper agreement.
  }
  \label{tab:result_discussion2}
  \begin{tabular}[width=\linewidth]{c|c|c|c|c|c|c|c|c}
    \hline
    \multicolumn{1}{c|}{\multirow{3}{*}{\textbf{model}}} &
    \multicolumn{4}{c|}{\textbf{CCD}} & \multicolumn{4}{c}{\textbf{JDDC}} \\ 
    \cline{2-9}
    &\multicolumn{2}{c|}{\textbf{Relevance}} & \multicolumn{2}{c|}{\textbf{Fluency}}   &
    \multicolumn{2}{c|}{\textbf{Relevance}} & \multicolumn{2}{c}{\textbf{Fluency}}\\
     \cline{2-9}
    &  Avg & $\kappa$ &  Avg & $\kappa$ &  Avg & $\kappa$ &  Avg & $\kappa$  \\
    \hline
    LSTM\cite{hochreiter1997long}  
    & 0.54  & 0.48  & 0.93   & 0.61
    & 0.53  & 0.52  & 1.09   & 0.59\\
    ByteNet\cite{gehring2017convolutional}            
    & 0.63  & 0.62  & 1.01  & 0.71
    & 0.59  & 0.55 & 1.19   & 0.67\\
    ConvS2S \cite{kalchbrenner2016neural}    
    & 0.64  &  0.51 & 1.05  & 0.82
    & 0.67  &  0.71 & 1.13  & 0.56\\
    S2S+attention\cite{nallapati2016abstractive}  
    & 0.89  & 0.55  & 1.32 & 0.69
    & 0.88 &  0.48 &  1.26 & 0.57\\
    PGN\cite{see2017get}      
    &  1.06 &  0.64 & 1.47  & 0.72
    &  0.96 &  0.69 & 1.52  & 0.53\\
    Transformer\cite{vaswani2017attention}
    & 1.02  & 0.71  & 1.41  & 0.65
    & 0.83  & 0.56  & 1.42 & 0.73\\
    CCN(vertical-only)
    & 1.04  & 0.58  & 1.56  & 0.68
    & 0.95  & 0.59  & 1.54  & 0.66\\
    \hline
    CCN$_{top-1}$
    & 1.03  & 0.61  & 1.59  & 0.75
    & 0.97  & 0.68  & 1.69  & 0.61\\
    CCN$_{top-2}$
    & 1.10  & 0.59  & 1.64  & 0.72
    & $\mathbf{1.01}$  & 0.72  & $\mathbf{1.77}$  & 0.70\\
     CCN$_{top-3}$
    & $\mathbf{1.12}$  & 0.66  & $\mathbf{1.69}$  & 0.68
    & 0.98  & 0.63 & 1.73  &  0.67\\
  \hline
    \end{tabular}
    \vspace{-8pt}
\end{table*}


\subsection{Evaluation Metrics}
We adopt two evaluation methods to validate the proposed model: Automatic Evaluation and Human Evaluation.
\subsubsection{Automatic Evaluation}
To evaluate the effectiveness of the dialogue generated by $\textbf{CCN}$, we used ROUGE \cite{lin2003automatic} and BLEU \cite{papineni2002bleu} scores to compare different models. We report ROUGE-1, ROUGE-L and BLEU to compare the advantages and disadvantages of each model.
\subsubsection{Human Evaluation}
In order to ensure the rationality/correctness of the generated utterance, we also conducted human evaluation. We randomly selected $300$ samples from the test set. Then, we recruited five annotators\footnote{All annotators took basic annotation training before the experiment.} to judge the quality of generated utterance from two perspectives: \cite{ke2018generating,zhu2019retrieval}:
\begin{itemize}
\item \textbf{Relevance}: Generated utterance is logically relevant to the dialogue context and can provide meaningful information.
\item \textbf{Fluency}: Generated utterance is fluent and grammatical.
\end{itemize}

The information on these two aspects are independently evaluated. For each aspect, we set three levels with scores: $+2$, $+1$, $0$, in which $2$ stands for excellent: for relevance, closely related to historical dialogue and be meaningful; for fluency, it has strong readability without grammatical error. $1$ stands for good: for relevance, with some off-topic information; for fluency, sentence is readable, but with slight grammatical error. $0$ means poor: for relevance, the sentence is off-topic or meaningless; for fluency, sentence has poor readability or serious grammatical errors. 
Finally, we obtain the weighted average score and kappa ($\kappa$) of each model to compare the effect of the model.

\subsection{Training Details}

During the training process, we set the dimension of word embedding as $300$ and use word2vec to build the initialization word vector.
The dimention of role embedding is set to $100$ with random initialization.
The hidden size is set to $300$, we use $4$ layer Transformer, where the number of heads equals to $8$. The dropout probability is set to $0.8$. 
Based on these settings, we optimize the objective function with a learning rate of
$5e-4$. We perform the mini-batch gradient descent with a batch size of $64$. We set maximum utterance length as 40 in decoder during generation (the generated sentence might contain sub-utterances).
\section{Result discussion}
\begin{figure*}
    \centering
    \includegraphics[width=1\textwidth]{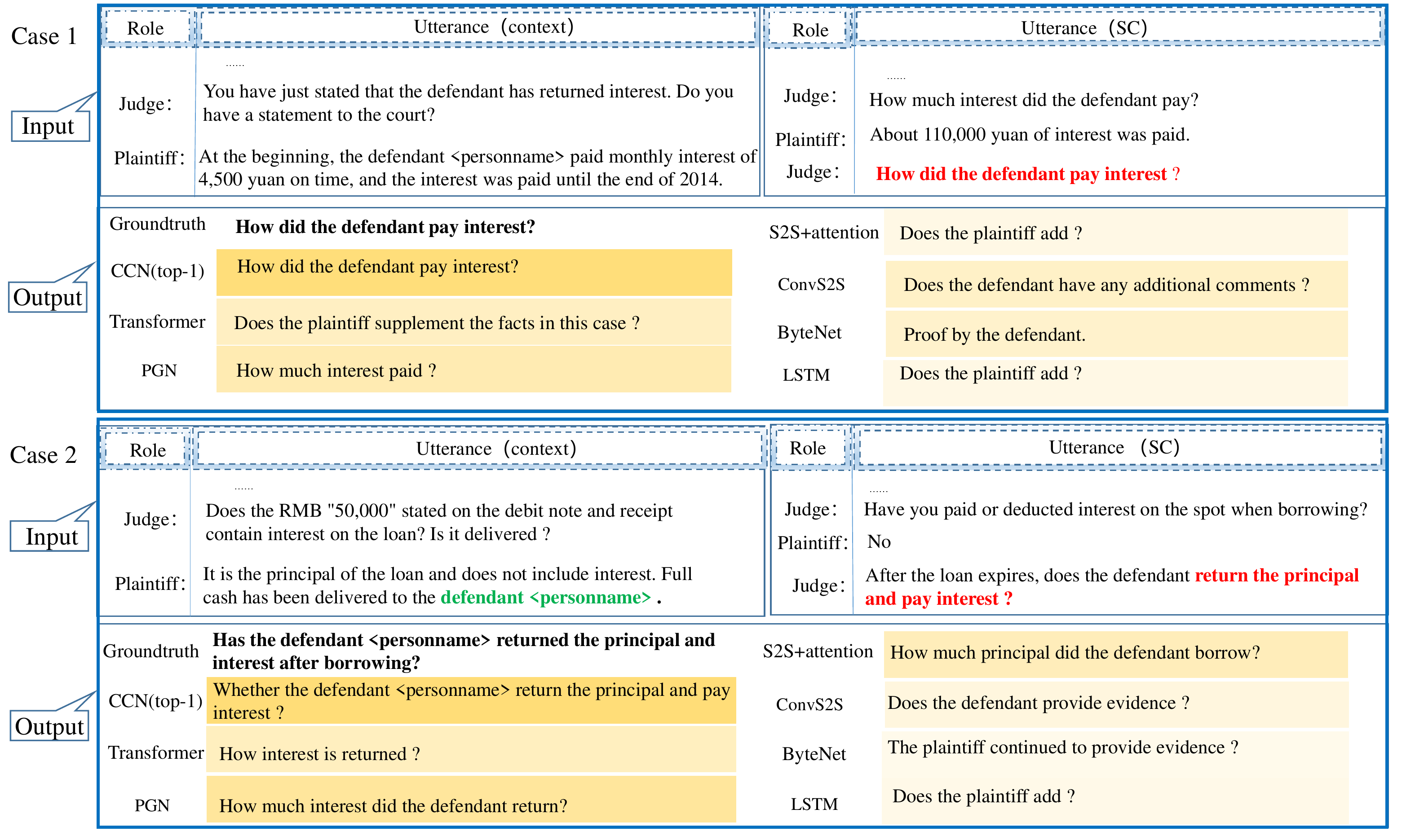}
    \caption{Case Study. We show two examples to intuitively illustrate the performance of all the tested models. The content in SC is the top 1 similar case used by our proposed method CCN. Note that the color indicates the BLEU score of each generated output. The darker the color, the higher the BLEU score is.}
    \vspace{-8pt}
    \label{fig:case_study}
\end{figure*}
\subsection{Overall Performance}
In the experiments, we select up to three similar cases to validate the effectiveness of the CCN, i.e., leveraging the most similar case \textbf{(top-1)}, top two similar ones \textbf{(top-2)}, and top three similar ones \textbf{(top-3)}. In addition, we also test the variant of \textbf{CCN(vertical-only)} by only adopting vertical copy from the context, which is similar to the setting of the baseline \textbf{PGN} but with the proposed hierarchical dialogue encoders. 

The performance of all the tested methods are reported in Table \ref{tab:result_discussion} and Table \ref{tab:result_discussion2} in terms of quantitative and qualitative evaluation, respectively. 
As Table \ref{tab:result_discussion} shows, the proposed approach \textbf{CCN} with its variants outperform all the baselines in Rouge and Bleu metrics over the two datasets. We can also observe the increasing performance as the number of referred similar cases increases. As for the two qualitative criteria, \textbf{CCN} also shows better performance by a big margin compared to the baselines. Note that the Kappa value ($\kappa$) indicates the agreement among the annotators.

As mentioned above, the increasing number of referred similar cases enables to bring about the improvement of performance, which demonstrates that the horizontal copy plays a critical role in dialogue generation without employing any external resources. However, in the training process, as the number of similar cases increases, the training speed is getting slower. Considering the time cost and memory limitation, only up to top three similar cases are utilized in this experiment to verify the proposed approach. 

\subsection{Case study}
Figure \ref{fig:case_study} shows two examples to illustrate the performance of different tested methods.
As depicted in case $1$, comparing with the baselines, the \textbf{CCN} can learn dialogue logic from \textbf{SC}s and accurately locate the sentence to complete the horizontal copy.
Another important finding is that we can use \textbf{SC}s to obtain more accurate representation information. It can identify specific entities from the context for vertical copy while capturing the discourse patterns from the similar cases for horizontal copy to finally synthesize the sentence to be generated. 

On the other hand, the baseline models are more inclined to generate general expressions which appear more frequent in the training data without much attention to the specific information in the context and the logical discourse patterns appearing in certain circumstances. 


\subsection{Error analysis}
In order to explore the algorithm limitation and model capability boundary, we summarize the samples with high error rate. The following observations should be highlighted to scope the limitation of current model and enlighten future investigation for this track of research. 

In the \textbf{CDD}, $53\%$ of errors\footnote{ the error refers to the generated text whose either relevance or fluency score equals 0.} occur when generated sentence contains the information that does not appear in context or in similar cases (e.g., “According to the provisions of Articles 44 and 45 of the Civil Procedure Law of the People's Republic of China, if the parties find that the members of the collegiate bench and the clerk experienced any of the following circumstances, they have the right to apply for their evasion orally or in writing.”). Similarly in \textbf{JDDC}, such problem caused $47\%$ of errors (e.g., “Sorry, we cannot refund you with the product [\#price, \#style, \#brand, \#specification, \#color] you required.”). Such kind of law/product related information might need to leverage external expert knowledge. In addition, $23\%$ and $36\%$ of errors occur in \textbf{CDD} and \textbf{JDDC} respectively, when it comes to the long sentence to be generated (e.g., the sentence length is more than $30$ words for the judge's inquiry or for customer service response). 

To address these problems in the future research, enhancing the long dependence of language models and establishing relations between different entities can be promising approaches.

\section{Related work}
\subsection{Pointer Network}
Pointer network \cite{vinyals2015pointer} is a special network structure. It solved the problem of generating sequence depending on the input sequence. 
Based on this basis, CopyNet \cite{gu2016incorporating} and PGN \cite{see2017get} were proposed, which can copy the words in the context and form output sequence to cope with Out-Of-Vocabulary ($\mathbf{OOV}$) problem.
Nowadays, pointer networks are increasingly popular in NLP applications. 
In text summarization,
\newcite{miao2016language} used it to select only suitable words from context instead of the entire dictionary for sentence compression; 
\newcite{nallapati2016abstractive} used it to speed up model convergence;
\newcite{sun2018multi} used it to generate text title;
\newcite{wang-etal-2019-concept} generated new conceptual words; 
\newcite{eric2017copy} used it to develop a recurrent neural dialogue system; 
\newcite{shen-etal-2019-improving} used it to enhance power of capturing richer latent alignment.
It was also widely used in many other tasks, such as
dependency parsing \cite{fernandez2019left, liu-etal-2019-hierarchical},
question answering \cite{kadlec2016text, golchha2019courteously}, 
machine reading comprehension \cite{wang2017machine, wang2017gated},
machine translation \cite{gulcehre2016pointing},
and language models \cite{merity2016pointer}.

Unlike previous studies, on the basis of internal copy, we introduce external copy, to establish a cross-copy structure, and achieve significant improvement.

\subsection{Dialogue System}

As an important task of NLP, dialogue system, has achieved great success and is widely used in practical applications, including customer service systems and chatbots. In recent years, with the development of deep learning technology, the neural network model has made significant progress:
\newcite{suimproving} solved the problem of information omitting and quoting in multiple rounds of dialogue by rewriting sentences;
\newcite{luconstructing} solved the problem of selecting the reply sentence in the dialogue system by adding features of time sequence and space;
\newcite{duboosting} solved the problem of lack of diversity in replies by using the dichotomy function to judge whether the two responses are similar.

There exist a number of prior studies to assist the task of dialogue generation through external knowledge:
\newcite{wu2019proactive, liu2018knowledge} leveraged the structured knowledge triples to assist dialogue generation.
Similarly, \newcite{li2019incremental, rajpurkar2018know, huang2018flowqa, reddy2019coqa} used documents as knowledge discovery for dialogue generation, and
\newcite{xia2017deliberation, ye2020knowledge, ghazvininejad2018knowledge, parthasarathi2018extending} utilized unstructured knowledge to explore in the open-domain dialogue generation.

With the deepening of dialogue generation, various new tasks have been proposed:
\newcite{lemultimodal} generated the most appropriate response by given video content, video title, and existing dialogue sentences; \newcite{tangtarget} introduced how to lead the conversation to a specific goal in an open conversation; \newcite{wangpersuasion} introduced how to use different persuasion strategies in the dialogue to persuade people to donate to charities; \newcite{caoobserving} were concerned about the application of dialogue analysis in the psychotherapy.

In our model, the \textbf{CCN} approaches to solve the problem of defective domain adaptation without any costly external knowledge.

\section{Conclusion and Outlook}
In this paper, we proposed a novel neural network structure-Cross Copy Networks, enabling both vertical copy (from dialogue context) and horizontal copy (from similar cases). Unlike prior models, the proposed CCN doesn't need additional knowledge input, and it can be easily adopted to other domains.
We conduct experiments on two different datasets with both quantitative and human evaluation to validate the proposed model. 
Experimental results proved CCN's superiority when comparing with a number of existing state-of-art text generation models, which tells the cross copy mechanism can successfully enhance the dialogue generation performance.

In future work, we will further investigate other content generation problems by leveraging multi-granularity copying mechanism. This study serves as the methodological foundation. 

\section*{Acknowledgments}
This work is supported by National Key R\&D Program of China (2018YFC0830200; 2018YFC0830206; 2018YFC0830700).

\bibliographystyle{acl_natbib}

\end{document}